\documentclass[11pt]{article}
\pdfoutput=1
\usepackage[final]{pdfpages}
\usepackage{fancyhdr}
\usepackage{textcomp}
\usepackage{longtable}

\usepackage{geometry}
\usepackage{titletoc}
\titlecontents{subsubsection}[2pt]{\addvspace{10pt}\bfseries\titlerule[0.5pt]\filright}{}{}{}[]
\titlecontents{section}[0pt]{\addvspace{5pt}\filright}{}{}{\dotfill\contentspage}[]
\titlecontents{subsection}[10pt]{\addvspace{1pt}\itshape\filright}{}{}{}[]

\newcommand{\tocTitle}[2]{\contentsline{section}{#1}{#2}{}\nopagebreak[4]}
\newcommand{\tocAuthors}[1]{\contentsline{subsection}{#1}{}{}}

\fancypagestyle{header}{
\lhead{\tiny  \hspace{-0.7in} \textbf{Proceedings of the 1$^{\rm{st}}$ International Workshop on Robot Learning and Planning (RLP 2016)} \\
\hspace{-0.7in} \textbf{in conjunction with 2016 Robotics: Science and Systems}\\
\hspace{-0.7in} \textbf{June 18, 2016 -- Ann Arbor, Michigan, USA}}
\lfoot{\tiny \hspace{-0.6in} \textbf{\textcopyright 2016 RLP}}
\cfoot{\thepage}
}

\fancypagestyle{rest}{
\cfoot{\thepage}
}

\begin{document}
\pagenumbering{roman}
\setcounter{page}{1}
\title{\bf Proceedings of the 1$^{\rm{st}}$ International Workshop on Robot Learning and Planning \\ (RLP 2016)}

\date{June 18, 2016}

\maketitle

\section*{Table of Contents}
  \tocTitle{Planning Non-Entangling Paths for Tethered Underwater Robots Using Simulated Annealing}{1}
  \tocAuthors{Seth McCammon and Geoff Hollinger}
  \tocTitle{Robust Reinforcement Learning with Relevance Vector Machines}{5}
  \tocAuthors{Minwoo Lee and Charles Anderson}
  \tocTitle{Path Planning Based on Closed-Form Characterization of Collision-Free Configuration-Spaces for Ellipsoidal Bodies, Obstacles, and Environments}{13}
  \tocAuthors{Yan Yan, Qianli Ma, and Gregory Chirikjian}
  \tocTitle{Learning the Problem-Optimum Map: Analysis and Application to Global Optimization in Robotics}{20}
  \tocAuthors{Kris Hauser}
  \tocTitle{Experience-driven Predictive Control}{29}
  \tocAuthors{Vishnu Desaraju and Nathan Michael}
  \tocTitle{The Impact of Approximate Methods on Local Learning in Motion Planning}{38}
  \tocAuthors{Diane Uwacu, Chinwe Ekenna, Shawna Thomas and Nancy Amato}

\pagebreak
\section*{Organizers}
\noindent
\begin{longtable}{p{0.25\textwidth}p{0.65\textwidth}}
Reza Iraji & Colorado State University \\
Hamidreza Chitsaz & Colorado State University \\
\end{longtable}
\section*{Program Committee}
\noindent
\begin{longtable}{p{0.25\textwidth}p{0.65\textwidth}}
Ali Agha & Qualcomm Research\\
Chuck Anderson & Colorado State University\\
Kostas Bekris & Rutgers University \\
Maren Bennewitz & University of Freiburg \\
Gianni Di Caro & Istituto Dalle Molle di Studi sull'Intelligenza Artificiale \\
Stefano Carpin & University of California-Merced \\
Hamidreza Chitsaz & Colorado State University \\
Howie Choset & Carnegie Mellon University \\
Juan Cort\'es & LAAS-CNRS \\
Angel P. del Pobil & Jaume I University \\
Thierry Fraichard & INRIA \\
Roland Geraerts & University of Utrecht \\
Kamal Gupta & Simon Fraser University \\
Adele Howe & Colorado State University \\
Seth Hutchinson & University of Illinois at Urbana-Champaign \\
Miles J. Johnson & Toyota Technical Center \\
Marcelo Kallmann & University of California, Merced \\
Lydia Kavraki & Rice University \\
Sven Koenig & University of Southern California \\
Jyh-Ming Lien & George Mason University \\
Dinesh Manocha & University of North Carolina at Chapel Hill \\
Rafael Murrieta-Cid & Center for Mathematical Research \\
Giuseppe Oriolo & Sapienza University of Rome \\
Wheeler Ruml & University of New Hampshire \\
Surya P. N. Singh & University of Queensland \\
Frank van der Stappen & Utrecht University \\
Chee Yap & New York University \\
\end{longtable}

\pagebreak

\section*{Best Paper Award}
\noindent
Learning the Problem-Optimum Map: Analysis and Application to Global Optimization in Robotics

\pagebreak

\pagenumbering{arabic}
\setcounter{page}{1}

\includepdf[fitpaper=true,pages=1,pagecommand={\thispagestyle{header}}]{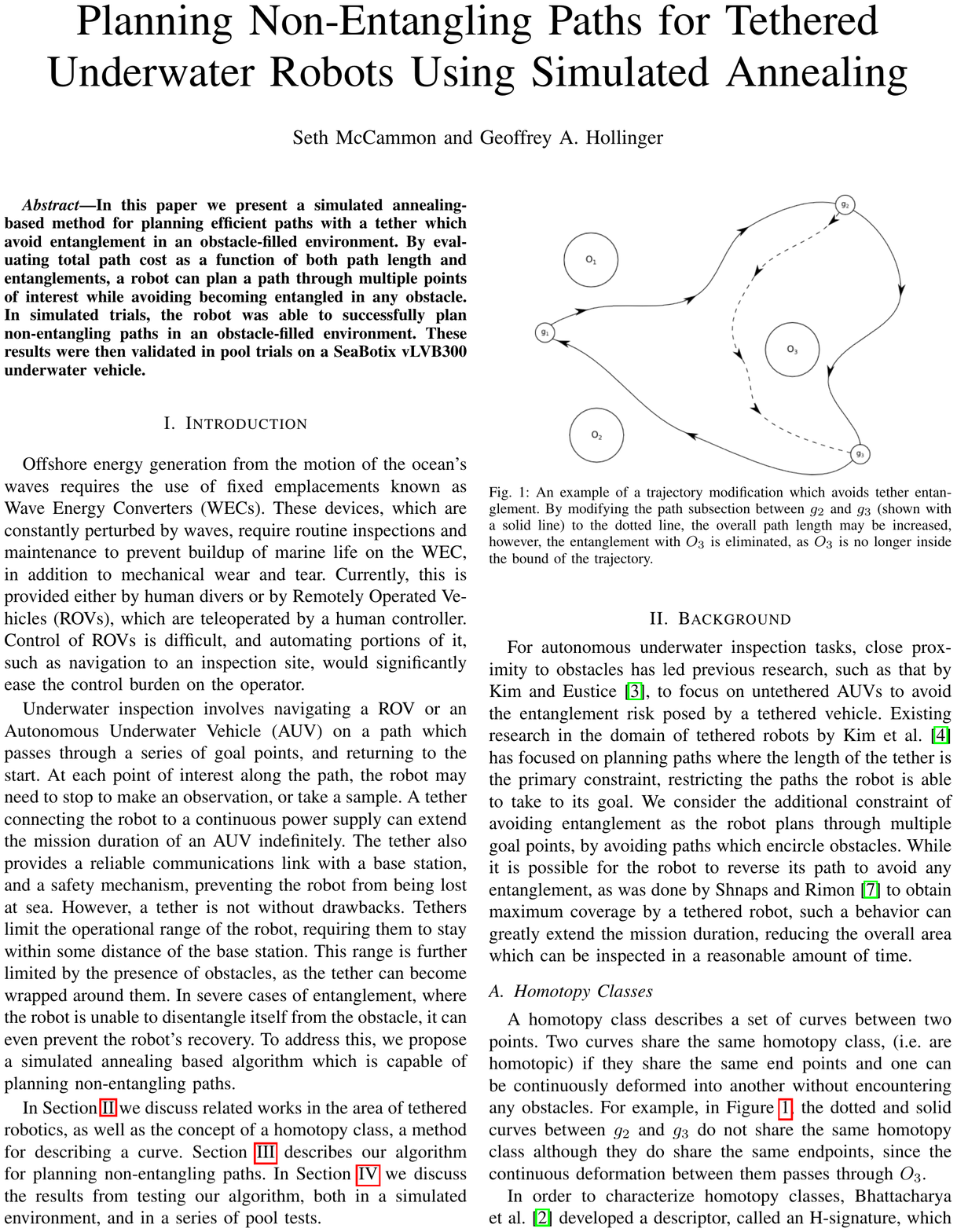}
\includepdf[fitpaper=true,pages=2-,pagecommand={\thispagestyle{rest}}]{RLP_2016_paper_2.pdf}
\includepdf[fitpaper=true,pages=1,pagecommand={\thispagestyle{header}}]{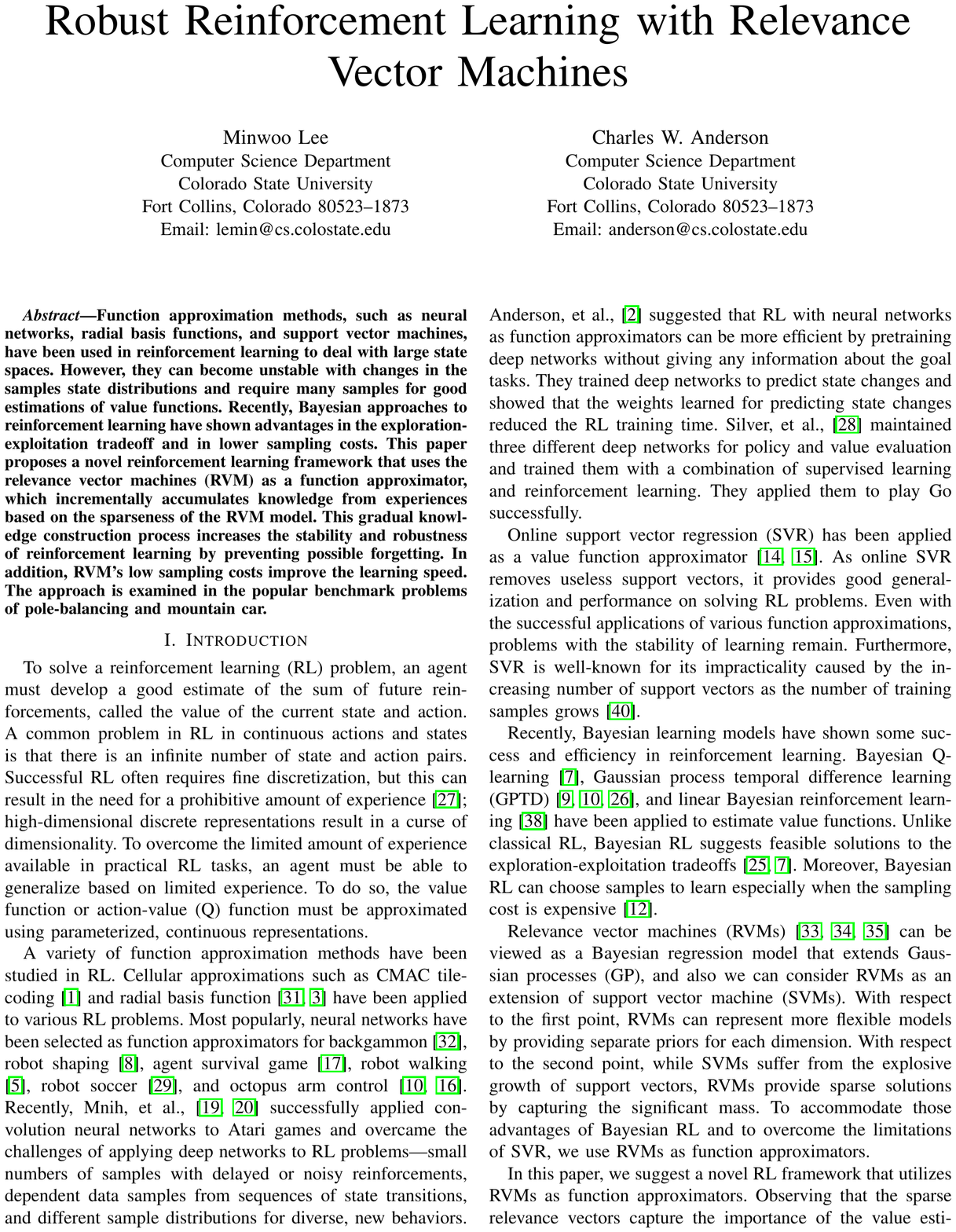}
\includepdf[fitpaper=true,pages=2-,pagecommand={\thispagestyle{rest}}]{RLP_2016_paper_4.pdf}
\includepdf[fitpaper=true,pages=1,pagecommand={\thispagestyle{header}}]{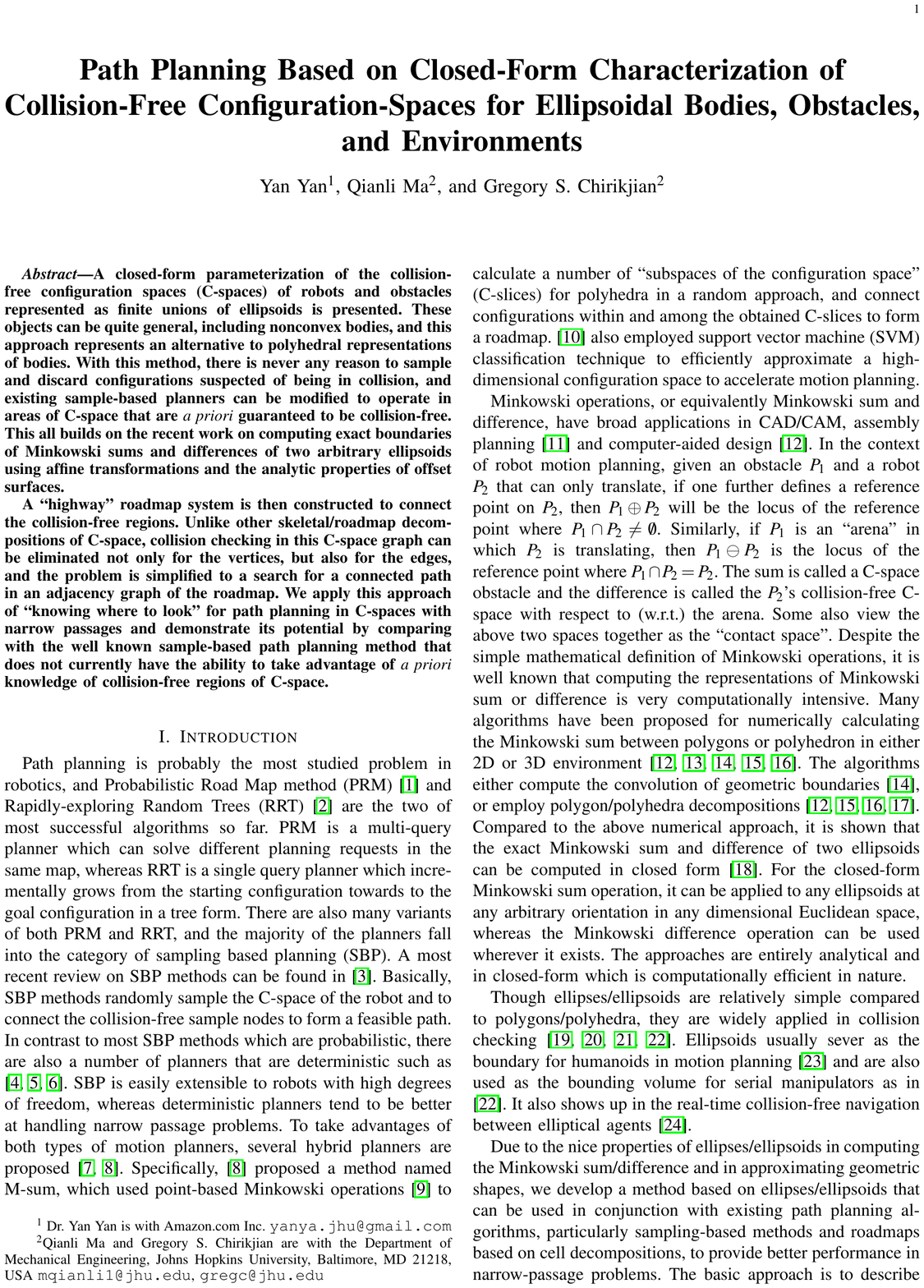}
\includepdf[fitpaper=true,pages=2-,pagecommand={\thispagestyle{rest}}]{RLP_2016_paper_7.pdf}
\includepdf[fitpaper=true,pages=1,pagecommand={\thispagestyle{header}}]{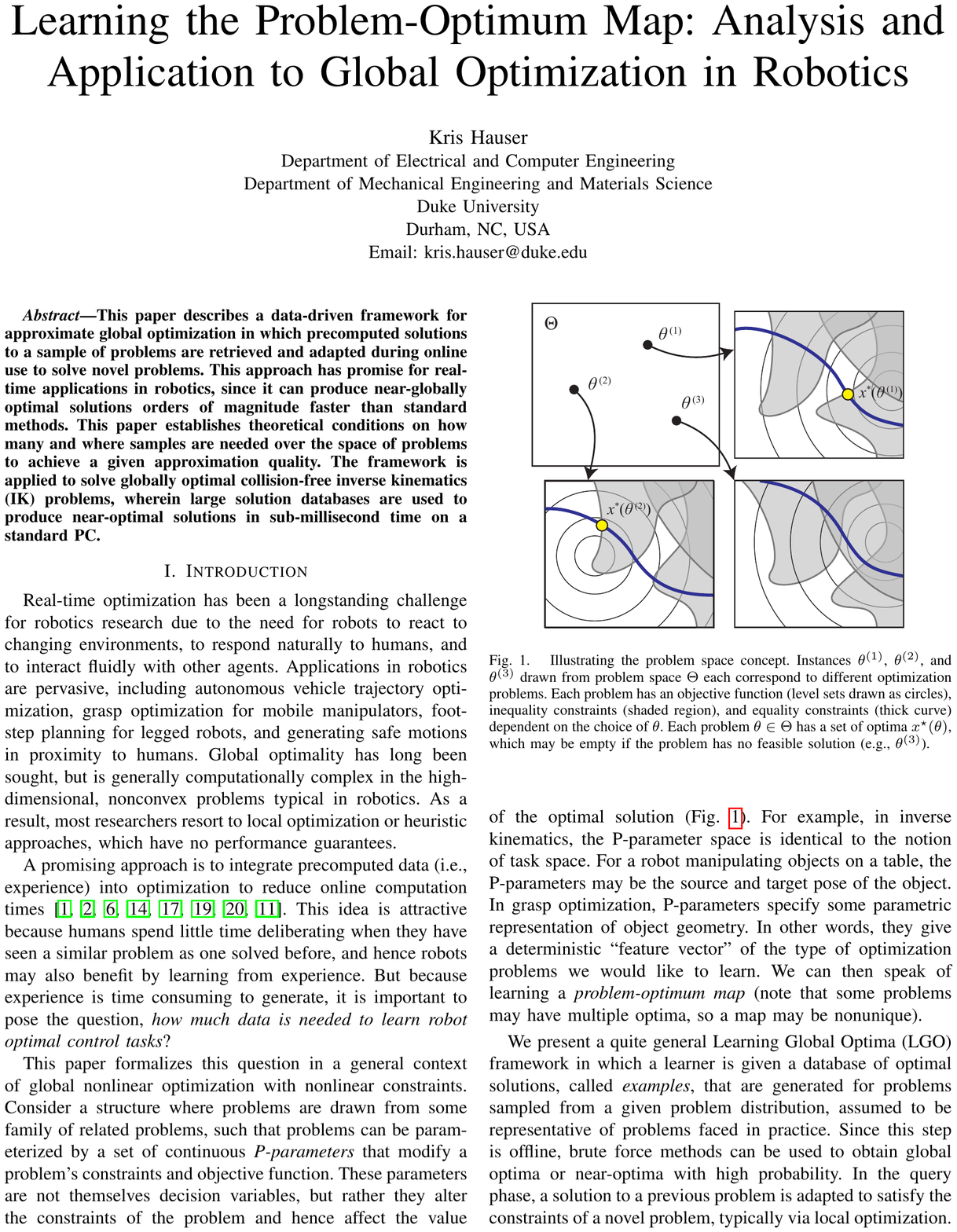}
\includepdf[fitpaper=true,pages=2-,pagecommand={\thispagestyle{rest}}]{RLP_2016_paper_8.pdf}
\includepdf[fitpaper=true,pages=1,pagecommand={\thispagestyle{header}}]{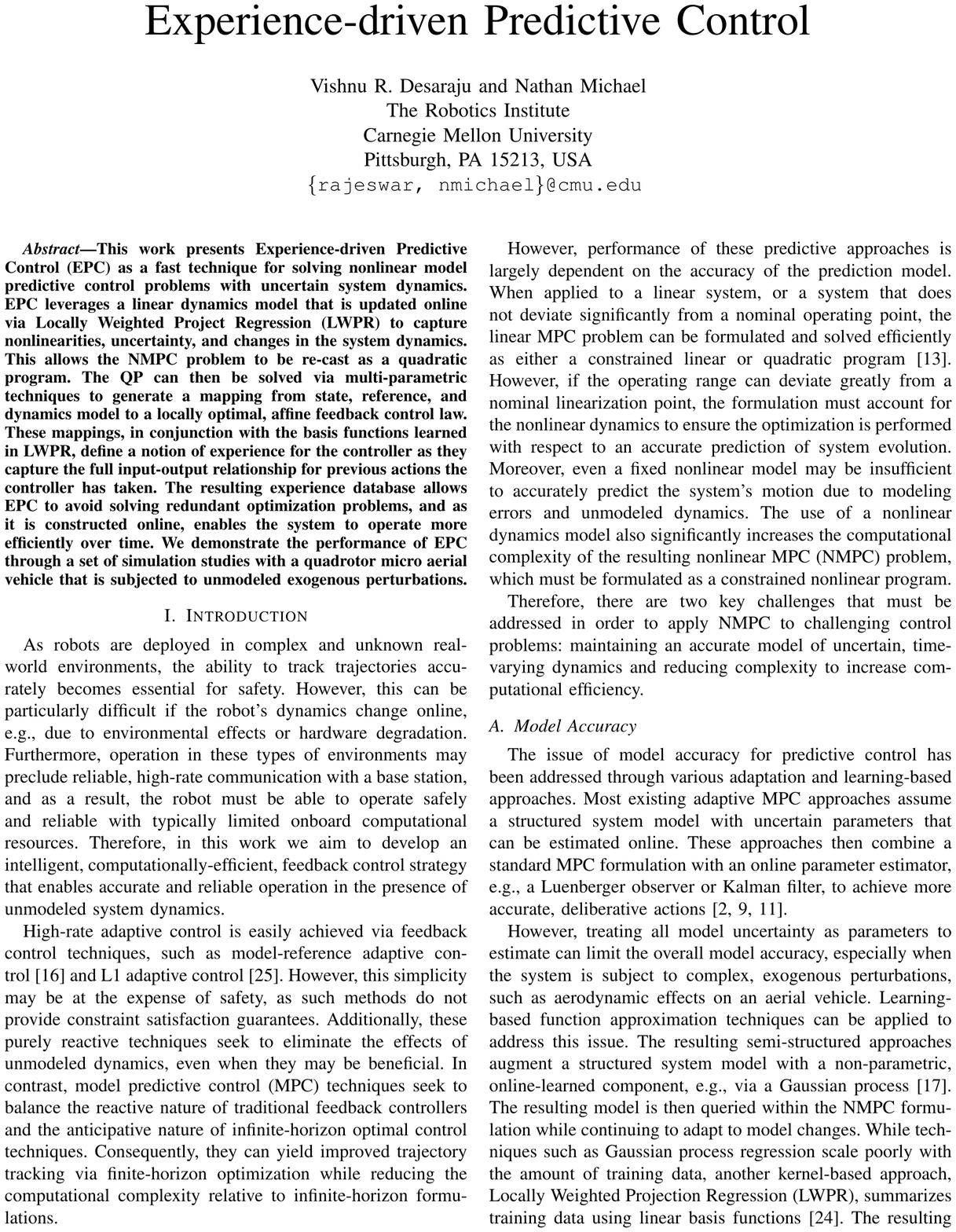}
\includepdf[fitpaper=true,pages=2-,pagecommand={\thispagestyle{rest}}]{RLP_2016_paper_10.pdf}
\includepdf[fitpaper=true,pages=1,pagecommand={\thispagestyle{header}}]{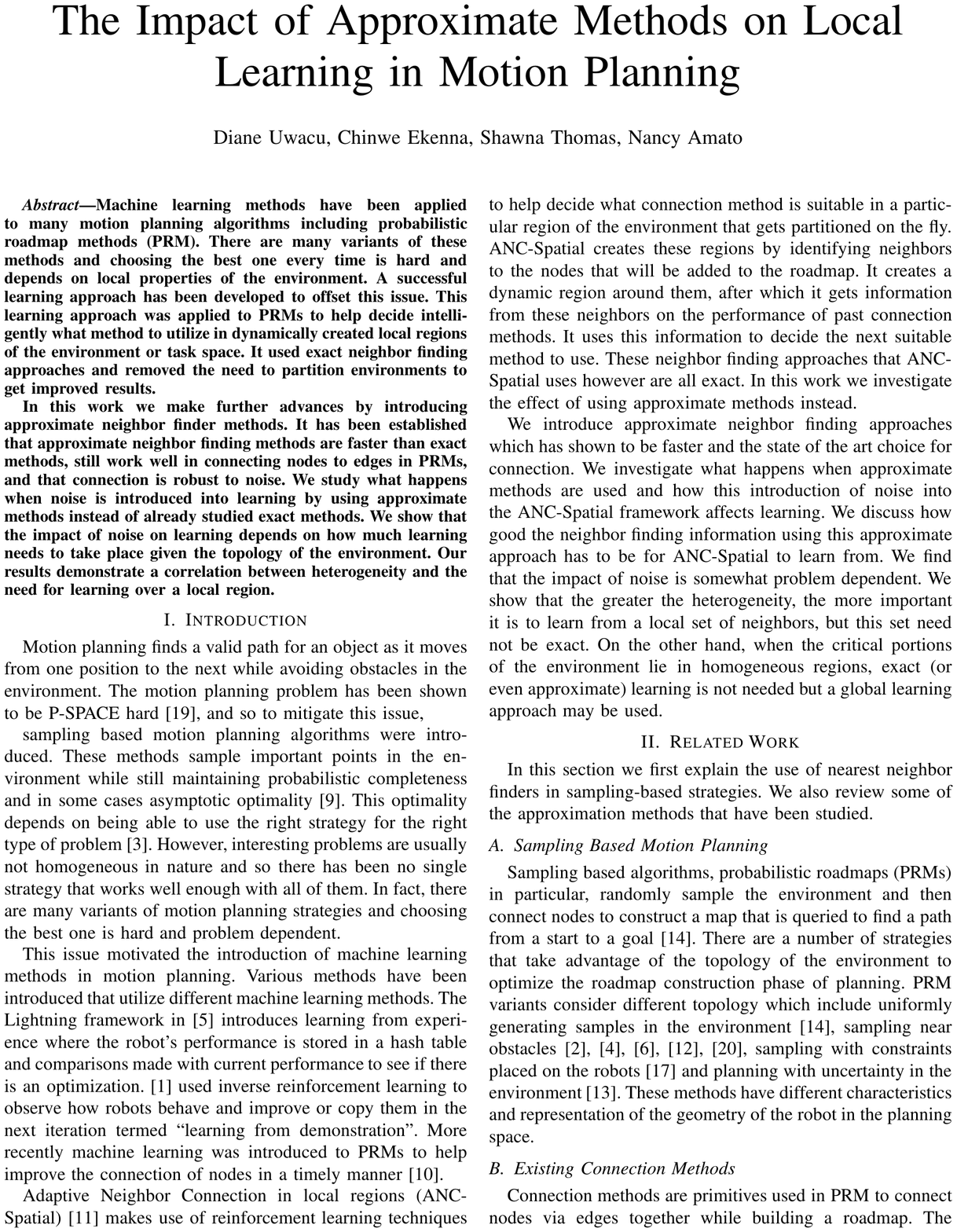}
\includepdf[fitpaper=true,pages=2-,pagecommand={\thispagestyle{rest}}]{RLP_2016_paper_11.pdf}
\end{document}